\documentclass[11pt]{article}
\pdfoutput=1

\usepackage[preprint]{acl}

\usepackage{times}
\usepackage{latexsym}

\usepackage[T1]{fontenc}

\usepackage[utf8]{inputenc}

\usepackage{microtype}

\usepackage{inconsolata}

\usepackage{graphicx}
\usepackage{inconsolata}
\usepackage{soul}
\usepackage{booktabs}
\usepackage{colortbl}
\usepackage{arydshln}
\usepackage[most]{tcolorbox}
\usepackage{subcaption}
\usepackage{caption}
\usepackage{placeins}
\usepackage{fontawesome5}
\usepackage{xspace}
\usepackage{multirow}
\usepackage{makecell}
\usepackage{authblk}
\usepackage{enumitem}

\definecolor{lightpink}{HTML}{f3e1f9}
\definecolor{lightpurple}{HTML}{e1e1f9}
\definecolor{lightblue}{HTML}{c6e8eb}

\definecolor{lightgreen}{HTML}{e1f9e1}
\definecolor{lightred}{HTML}{f9e1e1}
\definecolor{mustard}{HTML}{fff0c0}
\definecolor{lightgray}{HTML}{e3e2e0}

\DeclareRobustCommand{\gr}[1]{\sethlcolor{lightgray}\hl{#1}}
\DeclareRobustCommand{\green}[1]{\sethlcolor{lightgreen}\hl{#1}}
\DeclareRobustCommand{\red}[1]{\sethlcolor{lightred}\hl{#1}}

\newcommand{\simsmall}[0]{{\raise.17ex\hbox{$\scriptstyle\sim$}}
}

\usepackage[most]{tcolorbox}
\newif\ifcolorversion
\colorversionfalse   

\usepackage{xcolor}
\definecolor{revisioncolor}{rgb}{0.2,0.3,0.6} 

\DeclareRobustCommand{\revision}[1]{%
  \ifcolorversion
    \textcolor{revisioncolor}{#1}%
  \else
    #1%
  \fi}

\title{Debatable Intelligence: \\ Benchmarking LLM Judges via Debate Speech Evaluation}

\author[1]{\textbf{Noy Sternlicht}}
\author[2]{\textbf{Ariel Gera}}
\author[2]{\textbf{Roy Bar-Haim}}
\author[1,3]{\textbf{Tom Hope}}
\author[2]{\textbf{Noam Slonim}}

\affil[1]{\small School of Computer Science and Engineering, The Hebrew University of Jerusalem}
\affil[2]{\small IBM Research}
\affil[3]{\small The Allen Institute for AI (AI2)}
\affil[ ]{\small \faGlobe\xspace\href{https://noy-sternlicht.github.io/Debatable-Intelligence-Web}{Project}\xspace\xspace\xspace\faGithub\xspace\href{https://github.com/noy-sternlicht/Debatable-Intelligence}{Code}}

\begin{document}
\maketitle
\begin{abstract}
We introduce \emph{Debate Speech Evaluation} as a novel and challenging benchmark for assessing LLM judges. 
Evaluating debate speeches requires a deep understanding of the speech at multiple levels, including argument strength and relevance, the coherence and organization of the speech, the appropriateness of its style and tone, and so on. This task involves a unique set of cognitive abilities that previously received limited attention in systematic LLM benchmarking.
To explore such skills, we leverage a dataset of over $600$ meticulously annotated debate speeches and present the first in-depth analysis of how state-of-the-art LLMs compare to human judges on this task. Our findings reveal a nuanced picture: while larger models can approximate individual human judgments in some 
respects, they differ substantially in their overall judgment behavior. We also investigate the ability of frontier LLMs to \emph{generate} persuasive, opinionated speeches, showing that models may perform at a human level on this task.
\end{abstract}

\section{Introduction}
One particularly promising application of language models is the LLM-as-a-Judge paradigm (LLMaJ), where powerful LLMs are used to evaluate responses generated by humans or other models~\cite{Zheng2023JudgingLW}. This approach offers a scalable alternative to the costly 
process of collecting human-written references or annotations. But as LLMaJ systems become more prevalent, it is essential to rigorously evaluate their performance across a broad range of challenging 
and cognitively demanding 
tasks.

\begin{figure}[!htb]
    \centering
    \includegraphics[width=0.95\columnwidth]{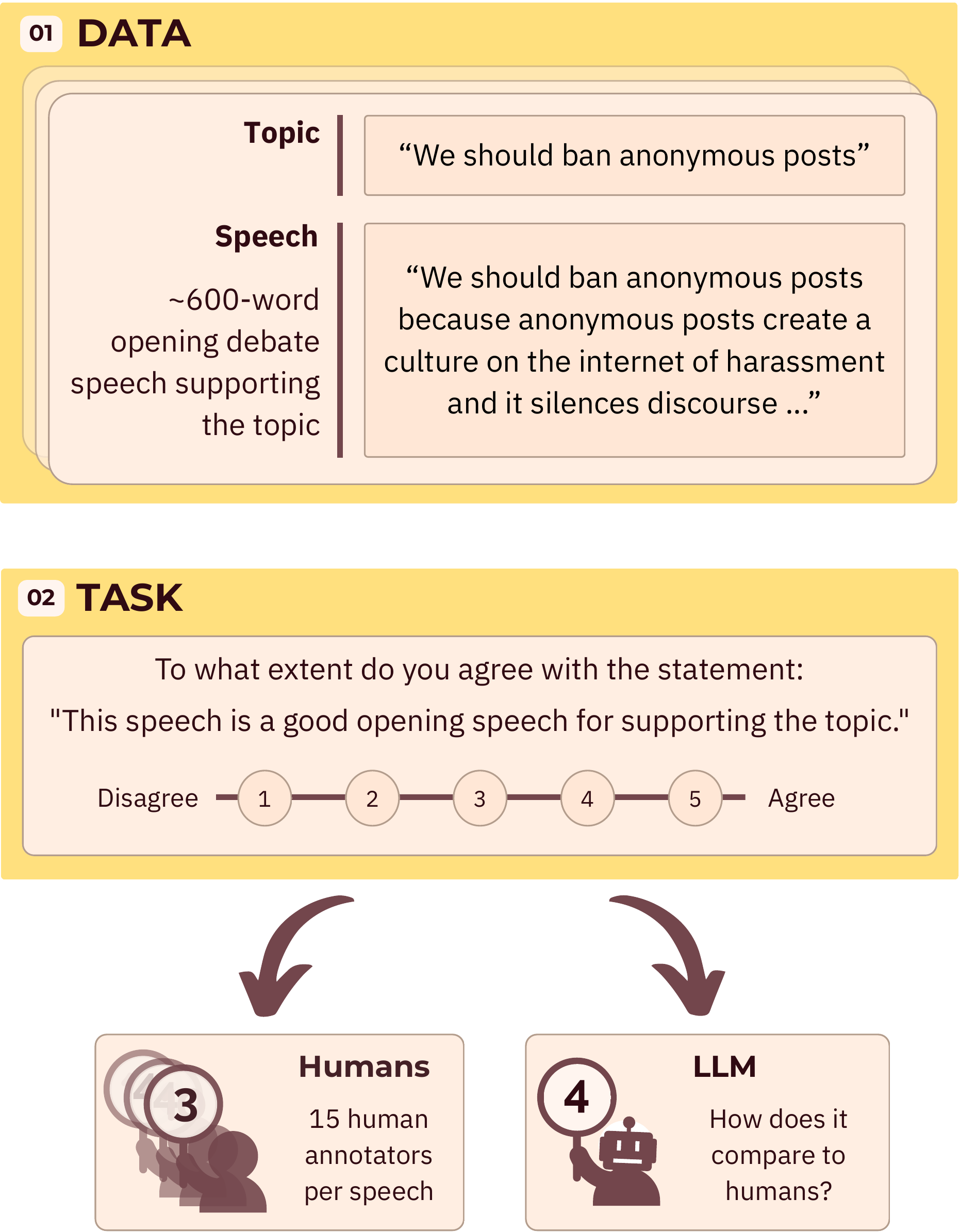}
    \caption{Benchmarking data (1) and task (2). We assess the judgment capabilities and behavior of LLMs by analyzing how they rate \emph{debate speeches} - long texts that argue for or against a controversial topic.}
    \label{fig:task_overview}
\end{figure}

Debate provides a rich, structured domain for this purpose. \emph{Evaluating} debate speeches is inherently complex and requires a nuanced assessment of multiple dimensions: persuasiveness, logical coherence, speech structure, and tone. In other words, the task demands a comprehensive understanding of long-form argumentation, making it a compelling testbed for LLMaJ systems.

In this work, we introduce a novel benchmarking task for LLM judges: 
evaluating long-form debate speeches that argue for or against controversial topics (Figure~\ref{fig:task_overview}).
This task involves unique cognitive challenges, such as comprehending the strength of the arguments and their relevance 
- abilities that remain unexplored by common judge benchmarking tasks like question-answering \cite{Zheng2023JudgingLW} or summarization \cite{wang2025dhpbenchmarkllmsgood}. We focus our investigation on the following questions:

\begin{itemize}[topsep=5pt, itemsep=2pt, parsep=2pt]
    \item \textbf{Q1}: How well do LLM judges perform the task of debate speech evaluation?
    \item \textbf{Q2}: In what ways does their judgment behavior differ from that of human judges?
\end{itemize}

We present the first comprehensive and systematic evaluation of LLM judges on debate speech rating. To assess this challenging and nuanced judgment task, we leverage a unique dataset of over $600$ speeches \cite{slonim2021autonomous}, carefully annotated by many human raters, yielding a high-quality benchmark for evaluating LLM judges.

Using this dataset, we conduct a series of experiments involving $23$ models of various sizes and types, including state-of-the-art reasoning LLMs. Our findings indicate that some LLMs align well with individual human annotators; size proves crucial, as models with fewer than 7B parameters consistently underperform. However, a closer analysis reveals that even the strongest LLM judges deviate from human judging behavior: while they often agree with humans on the relative ranking of speeches, they tend to assign systematically lower scores.

After measuring the performance and behavior of LLMs as judges, we pose an additional question:
\begin{itemize}[topsep=5pt, itemsep=2pt, parsep=2pt]
    \item \textbf{Q3}: How do \emph{speeches} by state-of-the-art LLMs compare to human speeches?
\end{itemize}

Our analysis reveals that judges rate speeches by GPT-4.1 higher than those of human expert debaters. This result demonstrates the impressive debating capabilities of modern LLMs. At the same time, it raises important social concerns regarding the potential misuse of LLMs.

To conclude, our contributions are as follows:
\begin{itemize}[topsep=5pt, itemsep=2pt, parsep=2pt]
\item We propose debate speech evaluation as a novel and challenging benchmark for the LLM-as-a-Judge paradigm.
\item We present the first comprehensive evaluation of LLM judging capabilities on this task, using complex, carefully annotated, debate data.
\item We analyze systematic differences between human and LLM judgments, and show that state-of-the-art models may surpass humans in composing argumentative text.
\end{itemize}

\section{Related Work}
\paragraph{LLM-as-a-judge}
The LLM-as-a-judge (LLMaJ) paradigm uses strong LLMs to assess or compare the output of other models or human-authored content. Recent works use it to aid or replace human-written references in various tasks. Some examples are evaluation of retrieval-augmented generation \cite{es2023ragasautomatedevaluationretrieval}, question-answering~\cite{Zheng2023JudgingLW, Asai2024OpenScholarSS} and writing ~\cite{Shao2024AssistingIW, Chiang2023CanLL}.

With LLMaJ becoming increasingly prevalent, properly evaluating the judges becomes crucial. Recent works attempt to benchmark LLMs judging capabilities in various areas. JudgeBench \cite{Tan2024JudgeBenchAB} is an LLMaJ benchmark that focuses on creating pairs of challenging responses, where the judge has to identify the correct answer. Another notable example is MM-Eval \cite{son2025mmevalmultilingualmetaevaluationbenchmark}, which focuses on multilingual tasks. Besides general benchmarking, many works are dedicated to analysing certain biases and issues associated with LLM judges, such as positional bias \cite{Wang2023LargeLM}, verbosity bias \cite{saito2023verbosity}, self-bias \cite{xu2024prideprejudicellmamplifies} or an excessive focus on style and grammar \cite{wu2023style}.

In our work, we explore the evaluation of long-form debate speeches, comparable in length to real-world texts such as opinion articles. This setting poses a unique challenge for LLMaJ, as it requires reasoning over both local properties, such as the strength and clarity of individual arguments, and more holistic dimensions, including the overall persuasiveness of the discourse. These aspects have not yet been comprehensively benchmarked in the context of LLMaJ.

\paragraph{Argumentative text evaluation}
Multiple works in the field of computational argumentation \cite{wachsmuth2017computational} have explored the quality of individual arguments \cite{Toledo2019AutomaticAQ, gretz2020large, durmus2024measuring, Habernal2016WhichAIA}. With the increased long-context capabilities of LLMs, researchers have begun exploring lengthy argumentative texts, applying LLMs to evaluate entire debates. A notable thread of research is using LLMs to determine the winner in a debate between two sides. \citet{Rescala2024CanLM} provide debater exchanges to a model and prompt it to predict the victor. \citet{Liang2024DebatrixMD} introduce a more sophisticated architecture that addresses long context length challenges in multi-round debates. Their model evaluates the debate speech by speech, using analysis of previous speeches rather than the entire debate history. \citet{Liu2024AnEA} analyze how several LLMs (GPT-3.5, GPT-4, and Llama-2-70B) predict debate winners and study their biases. 

Picking a debate winner, however, is a coarse measurement that hides much information and offers limited interpretability. Furthermore, existing debate datasets are either small in scale \cite{ruiz2021vivesdebate} or lack absolute quality annotations for individual participants \cite{Liang2024DebatrixMD, Durmus2018ExploringTR}. In this regard, the dataset introduced by \citet{slonim2021autonomous} stands out: each speech was assessed by multiple human annotators, yielding detailed and reliable quality judgments.

In this work, we are the first to use this richly annotated dataset as a benchmark to evaluate the judging capabilities of modern LLMs. To our knowledge, this constitutes the first large-scale benchmarking effort focused on quality scores for long debate speeches.

\section{Speech Quality Dataset}\label{sec:benchmark_desc}
In our work, we use the data collected by \citet{slonim2021autonomous} to benchmark LLM judges\footnote{\url{https://huggingface.co/datasets/ibm-research/debate_speeches}}. The data was initially collected to evaluate Project Debater, a system developed by IBM to participate in a competitive debate against humans. It includes hundreds of annotated debate speeches covering various topics (e.g., ``\textit{We should further exploit genetic engineering}''). The speeches were composed by Project Debater and various baselines, including human expert debaters and automated pipelines.

Each speech in the dataset was carefully rated by \revision{multiple crowd} annotators. Given the speech and the topic of the debate, the annotators were asked to score the following statement on a Likert scale between $1$ and $5$: ``\textit{This speech is a good opening speech to support the topic}'', as presented in Figure \ref{fig:task_overview}. \revision{To ensure annotation quality,} \citeauthor{slonim2021autonomous} further employed additional measures, such as ``control'' speeches added to identify unreliable annotators. \revision{This process yielded a total of $15$ trusted human annotations per speech. Appendix~\ref{sub:og_data_collection} provides further details on the annotation protocol and quality control measures used in \citet{slonim2021autonomous}.}

\revision{In this study, we analyze $631$ speeches (see Appendix~\ref{sub:our_data_partition}) drawn from the dataset introduced by \citet{slonim2021autonomous}. Table~\ref{tab:data_summary} outlines the main properties of the dataset.}

\begin{table}[!hbt]
\centering
\footnotesize
\begin{tabular}{lc}
\toprule
\multicolumn{2}{c}{\textbf{Speech Quality Dataset}} \\
\midrule
  \# Speeches   &  631 \\
  \# Human-written speeches  & 152 \\
  \# Synthetic speeches   & 479 \\
  \midrule
  \# Topics & 76 \\
  \# Speeches-per-topic (mean) & 8.3\\
  \midrule
  \# Annotators-per-speech & 15 \\
  \# Words-per-speech (mean) & 613.8\\
\bottomrule
\end{tabular}
\caption{Speech Quality Dataset statistics.}
\label{tab:data_summary}
\end{table}

\paragraph{Speech Sources} The data includes $152$ speeches transcribed from recordings of 
\textbf{human expert debaters}. In addition, it uses speeches generated by six synthetic pipelines: \textbf{(1) Summit} — based on extractive summarization \cite{feigenblat2017unsupervised}, \textbf{(2) Arg-Search} — using argument mining techniques \cite{stab2018argumentext}, \textbf{(3) Speech-GPT2} — generated with GPT-2 \cite{radford2019language}, \textbf{(4)-(5) Arg-Human1} and \textbf{Arg-Human2} — formed by concatenating human-written arguments, and finally \textbf{(6) Project Debater} — an automatic debate system.

The different sources exhibit varying levels of ``artificiality'': speeches of some sources (e.g., Speech-GPT2) are entirely artificial, while others are partly artificial (e.g., Arg-Human1,2), alongside $152$ completely human-authored speeches. This diversity gives the data a rich variety of writing styles and qualities. We provide additional details regarding the different sources in Appendix \ref{sub:og_data_collection}.

As the data was originally collected for \citet{slonim2021autonomous}, its construction relied on the most advanced models available at publication time (GPT-2). Therefore, it lacks annotations for current state-of-the-art LLMs. We complement this by presenting new experiments with GPT-4.1-generated speeches in Section~\ref{sec:new_speeches_eval}.

\section{Experimental Setup} \label{sec:experimental_settings}
We run a series of experiments to assess the performance of judges (\textbf{Q1}) and assess their behavior (\textbf{Q2}). These experiments involve running multiple LLMs as judges on the speech dataset 

and measuring their agreement with humans. 

Finally, we use GPT-4.1 to generate speeches, which we then evaluate with the highest-performing judges (\textbf{Q3}).

\begin{figure*}[!htb]
    \centering
    \begin{subfigure}[b]{\textwidth}
        \centering
        \includegraphics[width=0.9\textwidth]{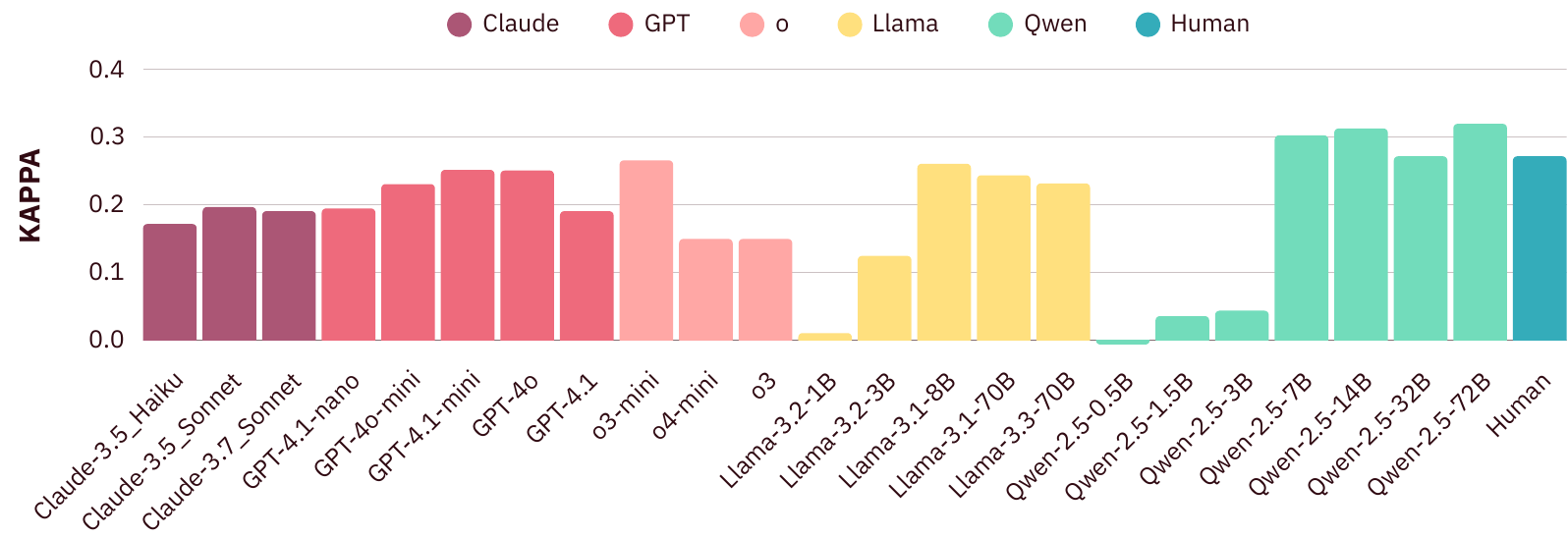}
        \caption{Average pair agreement}
        \label{fig:human_model_50}
    \end{subfigure}

    \vspace{1em}
    
    \begin{subfigure}[b]{\textwidth}
        \centering
        \includegraphics[width=0.9\textwidth]{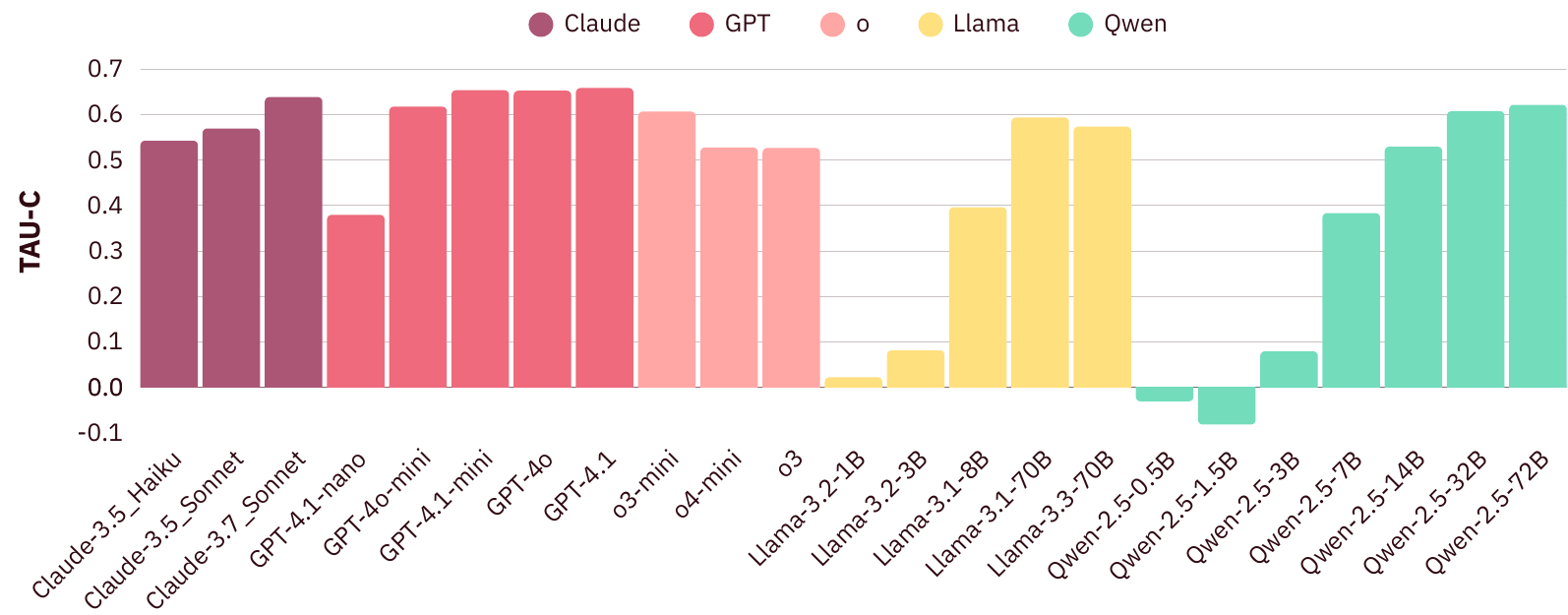}
        \caption{\revision{Ranking} agreement with average human scores}
        \label{fig:human_model_tau-c}
    \end{subfigure}

    \caption{
    We compare LLM-human agreement in two ways: Kappa scores (Figure \ref{fig:human_model_50}) and Kendall’s Tau with average human scores (Figure \ref{fig:human_model_tau-c}). Both show that larger models do better on this task, with a notable jump at \textbf{7B}.}\label{fig:human-model-agreement
    }\label{fig:human-model-agreement}
\end{figure*}

\paragraph{Judge models} Our experiments encompass $23$ models of different sizes, families, and licensing types (both proprietary and open), including state-of-the-art generative and reasoning models. We use various sizes of Qwen-2.5, Llama-3.1, Llama-3.2, and Llama-3.3. Additionally, we incorporate Claude-3.5, GPT-4o, and GPT-4.1 models. We also employ four reasoning models: Claude-3.7-Sonnet, o3, o3-mini, and o4-mini. Appendix \ref{sec:judge-implem-details} describes additional details regarding execution.

\paragraph{Prompts} The prompt for the judge models follows the instructions given to human annotators in \citet{slonim2021autonomous}. It asks the model to score input speeches on a scale of 1 to 5. We also create a variant of this prompt that requires judges to justify their scores. Prompts are provided in Appendix \ref{sec:judge-implem-details}.

\paragraph{Agreement metrics} We define the following metrics for measuring agreement on speech ratings. Following \citet{slonim2021autonomous}, we employ weighted Kappa agreement. For human annotators, we identify pairs who share at least $50$ speeches (termed the ``\textit{minimal-sample}'' threshold) and calculate their average pairwise agreement scores. \revision{This threshold ensures that agreement is estimated on a sufficiently robust sample, reducing the variance that would arise from very small overlaps.}

\revision{We adopt a leave-one-out evaluation setup, which is common in the LLM-as-a-Judge literature \cite{Zheng2023JudgingLW}. For each pair of human annotators who rated the same set of speeches, we replace one annotator with an LLM and compute agreement with the remaining human on the overlapping subset. We perform this substitution for both positions in each pair and average the pairwise agreements.}

\begin{figure*}[!htb]
    \centering
    \includegraphics[width=0.9\textwidth]{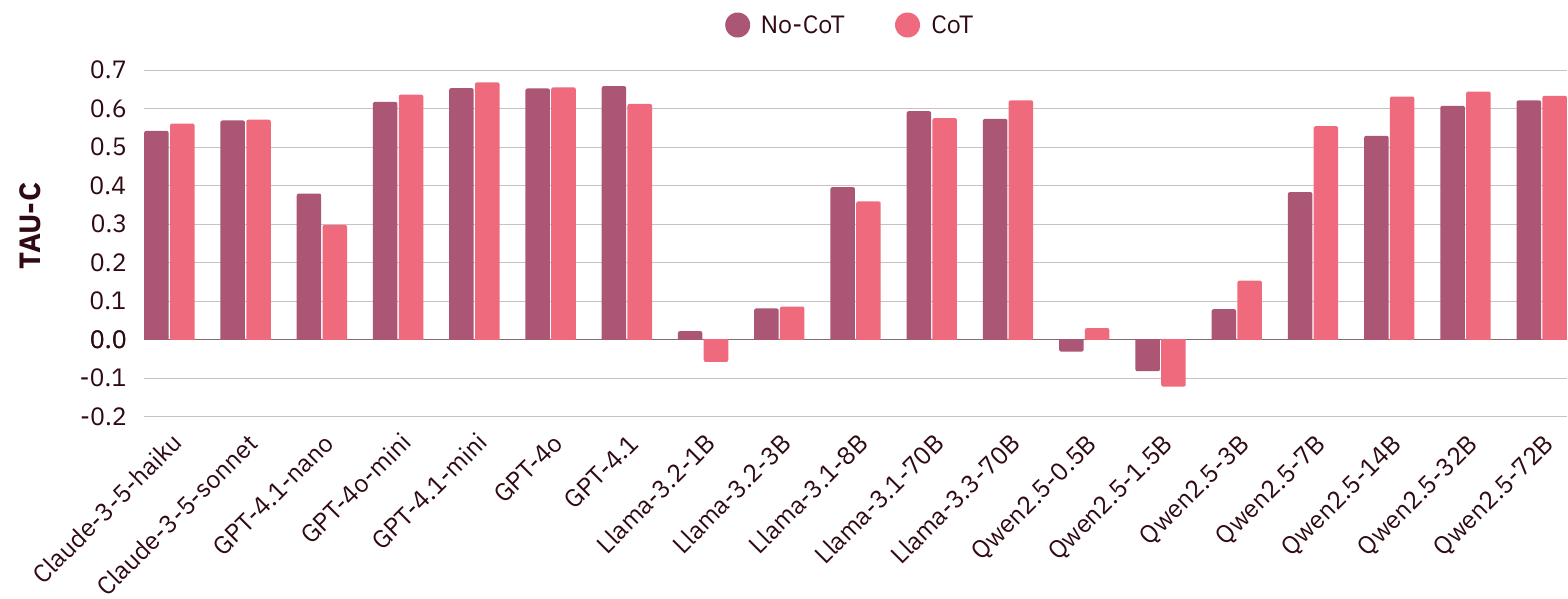}
    \caption{LLMs, especially stronger ones, generally align better with human annotators when using CoT prompting.}
    \label{fig:prompt-comp}
\end{figure*}

Even when enforcing a \textit{minimal-sample} value, the number of shared speeches per annotator pair remains \revision{limited}, which could introduce noise \revision{and leaves much of the dataset unused}. We therefore employ an additional, more robust metric that measures ranking correlation between an LLM judge and the \textit{average} of $15$ human annotations per speech, over the \emph{entire} set of speeches. We use the Kendall's Tau-C correlation metric as it can handle agreement between variables with different possible values -- in our case, a continuous variable (the average human scores) and a categorical one (the LLM judge scores).

\paragraph{Speech generation} We use GPT-4.1 to generate new speeches. We request the model to generate \revision{two} $600$-word speeches about a given debate topic \revision{(similarly to human-authored speeches)}. Appendix \ref{sec:speech_gen_implementation} shows the prompt used for this step, along with additional technical details.

\section{Results}
\subsection{Judge Performance Evaluation} \label{sec:human-model_agreement}

\paragraph{Humans vs. LLM judges} We start by analyzing the judges' performance, aiming to examine to what extent the judges are \emph{comparable} with human annotators, and can replace them on this task. Figure~\ref{fig:human_model_50} presents the average pairwise Kappa agreement between humans, along with results from repeating the experiment while replacing one annotator with various LLM judges (as described in Section \ref{sec:experimental_settings}). Our results indicate that there are LLM judges who perform on par with humans, with some, like Qwen-72B, even surpassing human performance \revision{--- meaning that their agreement with a human annotator is higher than human agreement.}

\paragraph{Are stronger models better?} Figure \ref{fig:human_model_50} shows that models larger than 7B are generally more capable on this task. However, smaller models such as Qwen-7B and Qwen-14B \revision{seem to} outperform considerably stronger and more advanced models like GPT-4o. To address potential noise from the limited number of speeches shared between annotator pairs, we introduce a more robust analysis. As described in Section \ref{sec:experimental_settings}, we compute the \revision{ranking} agreement between average human scores per speech and LLM judges using Kendall's Tau-C. \revision{This score represents the ability of a judge to reproduce the relative ranking of speeches given by the human annotators.}

\begin{figure}[!htb]
    \centering
    \includegraphics[width=\columnwidth]{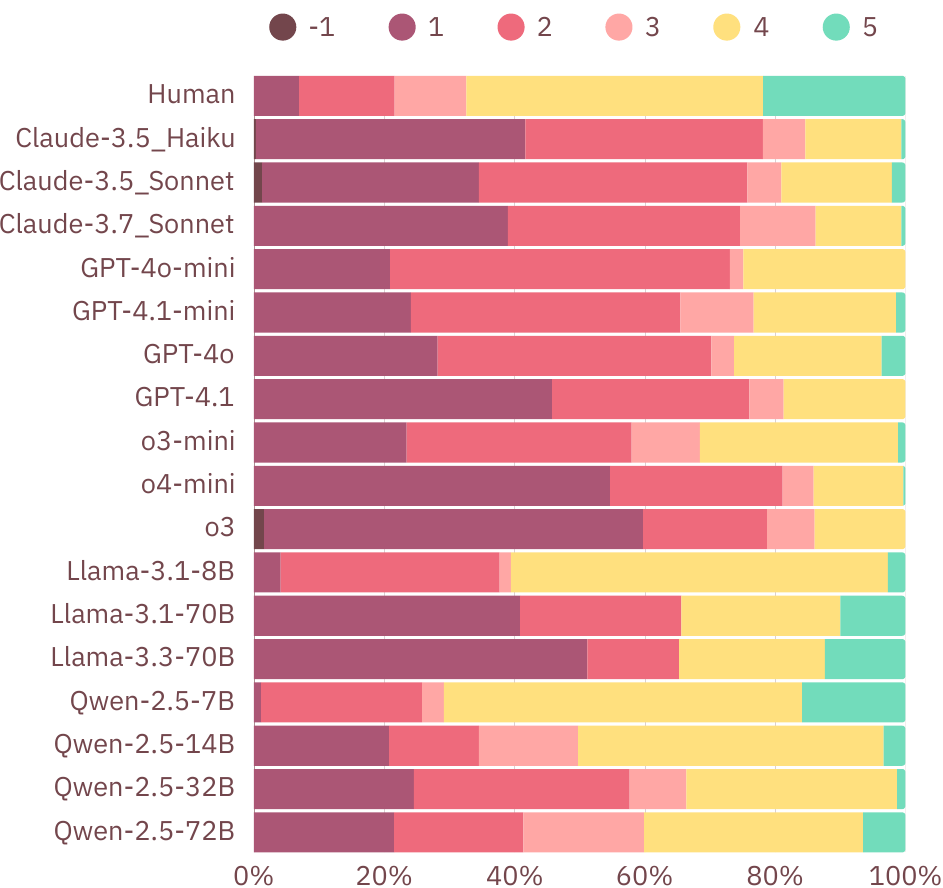}
    \caption{Strong LLM judges tend to give \textit{lower} scores than human annotators (-1 signifies parsing issues). Results for all judges are shown in Appendix \ref{sec:full_scores_dist}.}
    \label{fig:scores_dist}
\end{figure}

Figure \ref{fig:human_model_tau-c} shows the results of this experiment across the entire dataset. Like \ref{fig:human_model_50}, these additional findings confirm that larger LLMs generally show higher agreement with humans, aligning well with expectations based on model size and capability. Yet, we observe that some state-of-the-art models, such as o3, still underperform on this task.

\begin{figure*}[!htb]
    \centering
    \begin{subfigure}[b]{\textwidth}
        \centering
        \includegraphics[width=\textwidth]{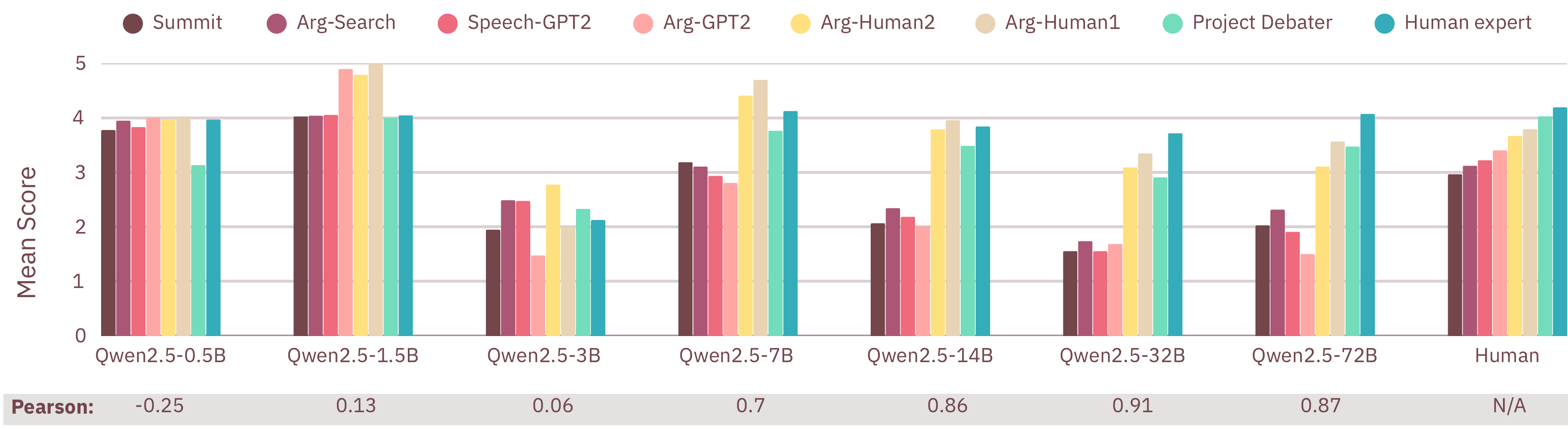}
        \caption{The effect of model size on rating speeches from different sources}
        \label{fig:source_ranking_size}
    \end{subfigure} 

    \vspace{1em}
    
    \begin{subfigure}[b]{\textwidth}
        \centering
        \includegraphics[width=\textwidth]{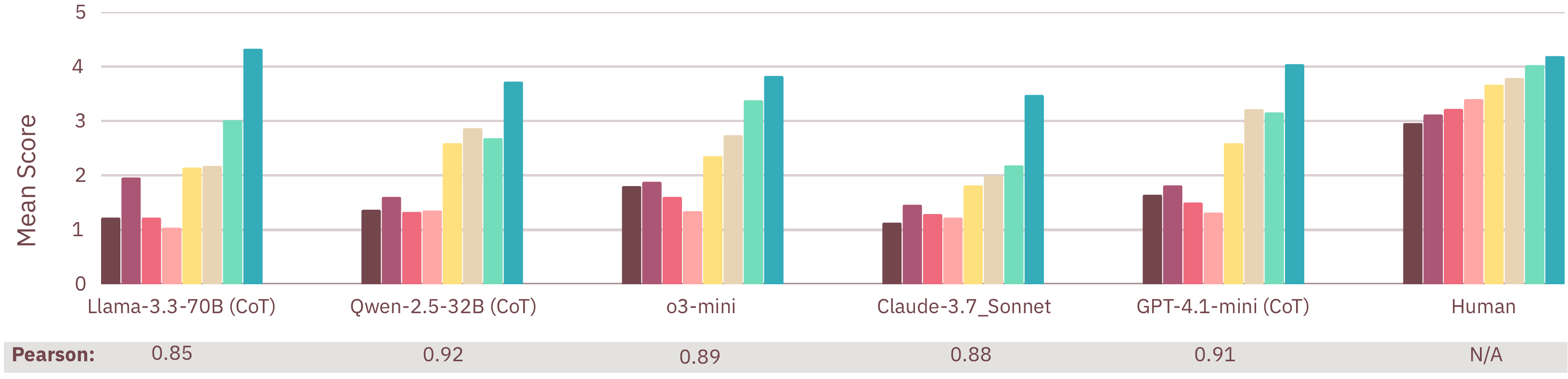}
        \caption{Strong judges' rating of speeches from different sources.}
        \label{fig:source_ranking_strong}
    \end{subfigure}
    \caption{Judge ratings of different speech sources, compared to human ratings (rightmost panels). \gr{Text in gray} shows the \textbf{Pearson correlation} \revision{between the mean scores per source given by humans and LLMs.}
    As models become more capable, they rank speeches from different sources more similarly to how humans do.  
    Interestingly, while stronger models align with humans on the relative ranking of speeches, they tend to assign lower overall scores to synthetic data.}\label{fig:source_analysis}
\end{figure*}

\paragraph{Enhancing zero-shot judges}
Inspired by recent works \cite{Liu2023GEvalNE, luo2023chatgpt}, we test whether Chain-of-Thought (CoT) prompting can help to improve the performance of LLM judges.
To this end, we repeat the previous experiment using a CoT variant of our prompt (available in Appendix \ref{sec:judge-implem-details}). Figure \ref{fig:prompt-comp} presents our results. For most larger models, CoT prompting offers a slight improvement. Interestingly, it degrades some of the smaller judges. We hypothesize that some performance decrease could be attributed to increased parsing issues (see Appendix \ref{sec:parsing_errors}).

We make an additional attempt to improve the LLM judges through ensembling, which we describe in Appendix \ref{sec:judge_ensem}.

\subsection{Judge Behavior Analysis} \label{sec:behavior}
\paragraph{Judge score distributions} Figure \ref{fig:scores_dist} presents the distribution of speech scores given by various LLM judges and human annotators. We observe that stronger LLMs (larger than 7B) generally give the speeches lower scores. This is interesting, given the results presented in Figure \ref{fig:human_model_tau-c}, which show that the same models have high agreement with humans on speech rankings. We conclude that LLM score distributions are not well aligned with humans, even when the LLMs capture a similar instance-level ranking to humans.

The score distribution also helps explain why some LLMs outperform others in one metric but not in another. We observe that models whose score distributions resemble those of humans generally obtain better Kappa scores, even if their agreement on the speech ranking is relatively low. For example, Llama-3.1-8B has a higher Kappa agreement than Llama-3.1-70B, but a lower Kendall's tau score. This could be attributed to Kappa normalizing agreement based on the expected agreement by chance, which is derived from the scores distribution. This is in contrast to Kendall's tau, which is a non-parametric measure of agreement.

\paragraph{Speech source analysis} \label{sec:source_analysis}
 Evaluating debate speech quality is a complex task, requiring attention to both local aspects (such as argument strength) and holistic elements like coherence and flow. As described in Section \ref{sec:benchmark_desc}, the benchmark data includes speeches from various sources with different levels of artificiality. Some sources contain entirely generated texts (like Speech-GPT2, which may have better flow but weaker arguments), while others automatically construct speeches using human-written arguments (which may have stronger arguments but are harder to follow). Studying modern LLM judges across these different sources could reveal insights into their judging behavior and which qualities they value most in effective debate speeches.

  \begin{figure}[!htb]
    \centering
    \includegraphics[width=\columnwidth]{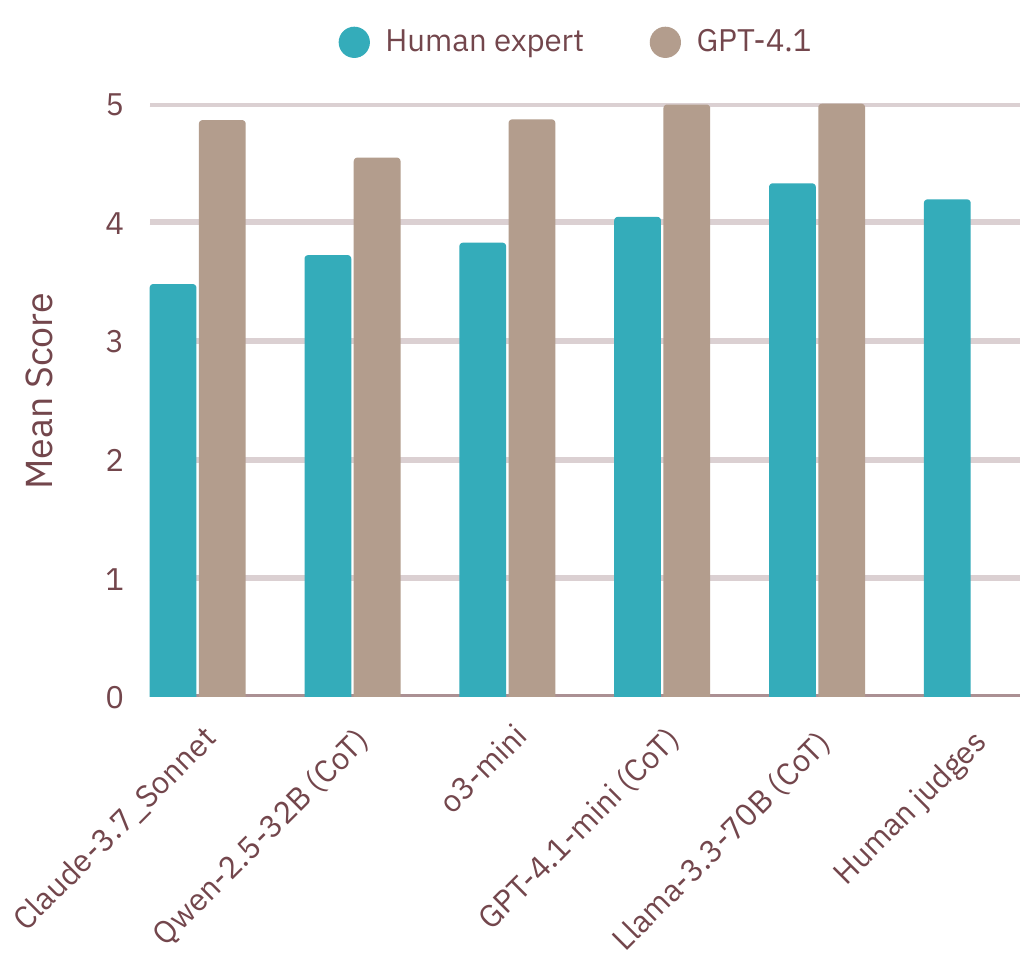}
    \caption{Strong judges rate speeches generated by GPT-4.1 \textit{higher} than those by human-expert debaters.}
    \label{fig:gpt_4_1_speeches}
\end{figure}

 In Figures \ref{fig:source_ranking_size} and \ref{fig:source_ranking_strong}, we present the average scores that LLMs and human judges assign to speeches from different sources. The rightmost panels present the average human scores, taken from \citet{slonim2021autonomous}. Figure \ref{fig:source_ranking_size} compares models from the Qwen family. We see that as models grow in size, their rating of speeches from different sources increasingly resembles the rating given by humans; we also again observe a turning point occurring around the 7B mark, as described in Section \ref{sec:human-model_agreement}.

In line with the general score distributions presented in Figure \ref{fig:scores_dist}, models that strongly align with humans (such as Qwen-72B) tend to assign lower scores. Interestingly, this is most prominent for speeches from synthetic sources: While Qwen-72B gives human speeches an average score of $\simsmall{4}$ (akin to human annotators), it gives artificial speeches (like speech-GPT2) a score of $\simsmall{1.5}$ (almost two points lower than human judges!). 
We also note that the score gap for partially artificial sources (like Arg-Human2) is not as dramatic. 
As shown in Figure \ref{fig:source_ranking_strong}, other strong judges exhibit similar scoring patterns.
We provide examples of speeches from different sources in Appendix \ref{sec:speech_examples}.

Our findings could be explained by stronger judges showing a more decisive behavior, amplifying the gap between strong and weak ``systems'' (in our case, groups of speeches; see \citealp{gera2024justrank}). Another factor that might contribute to the difference between human and LLM judges' behaviour is that, in contrast to LLMs, multiple humans annotate the same speech, smoothing out extremes and \revision{possible} biases of individual judges.

\subsection{Speech Quality of Modern LLMs}
\label{sec:new_speeches_eval}
Having established that certain LLM judges strongly agree with humans, we examine how the best-performing judges from each family (those with the highest Tau-C correlation) evaluate speeches generated by a SOTA model. We use GPT-4.1 to generate $152$ speeches (two per debate topic, similarly to speeches by humans) and compare their average rating to that of human speeches. The generation prompt is provided in Appendix \ref{sec:speech_gen_implementation}.

Our results, presented in Figure \ref{fig:gpt_4_1_speeches}, indicate a strong preference for GPT-4.1 speeches over those authored by humans. This is in stark contrast to the results in Section \ref{sec:source_analysis}, where older synthetic sources were given a very low rating.

All five models, including those not subject to self-bias like Claude-3.7 and Llama-3.3-70B, rated the generated speeches higher than the human-authored ones. While we lack human annotations for further verification, these results demonstrate the rapid progress of generative models in recent years. More importantly, the ability of these models to argue controversial topics \revision{-- ostensibly, in a more effective manner than human experts --}
raises concerns about their safe deployment and potential misuse. We provide an example of a GPT-4.1 generated speech in Appendix \ref{sec:speech_examples}, Figure \ref{fig:gpt_41_speech}.

\subsection{Judge Reasoning Analysis} \label{sec:reasoning_analysis}
In this section, we drill down into the 
differences in scores that LLM judges assign to speeches from various sources. 
To better understand these differences, we apply Key Point Analysis \cite{BarHaim2020FromAT, BarHaim2020QuantitativeAS} to chain-of-thought explanations generated by Llama-3.3-70B, one of the capable judges we examine in Sections \ref{sec:source_analysis} and \ref{sec:new_speeches_eval}. Key Point Analysis (KPA) is a summarization technique used to compress a corpus of texts into a group of concise \textit{pro} (positive) and \textit{con} (negative) key points. We use it to identify a canonical group of explanations or justifications the judge provides for its speech scores. Appendix \ref{sec:kpa_implementation} contains additional implementation details, along with similar results for two more judge models.

\begin{figure*}[!htbp]
    \centering
    \includegraphics[width=\textwidth]{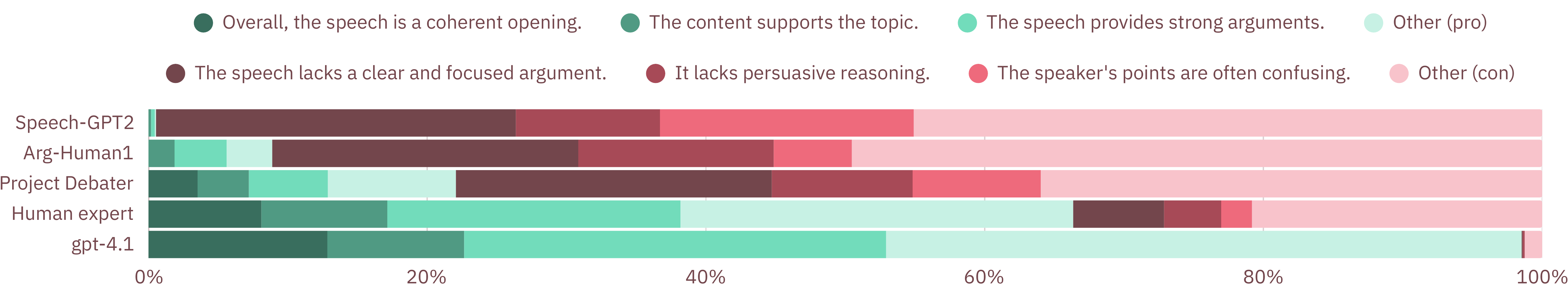}
    \caption{Distribution of the top three \green{pro} and \red{con} key points in \textbf{Llama-3.3-70B}’s chain-of-thought explanations, grouped by source. \green{Other (pro)} and \red{Other (con)} denote less frequent points. The relative share of positive key points per source reflects the judge ratings discussed in Sections~\ref{sec:source_analysis} and~\ref{sec:new_speeches_eval}.}
    \label{fig:llama-kpa}
\end{figure*}

Figure \ref{fig:llama-kpa} shows the distribution of the three most common pro and con explanations across speech sources. We first note that the proportion of positive key points per source aligns with the judge's rating in Sections \ref{sec:source_analysis} and \ref{sec:new_speeches_eval}.  Additionally, we see that some key points focus on more holistic aspects of the speech, such as persuasiveness, relevance, and coherence (e.g., ``\textit{Overall, the speech is a coherent opening}''), while others, like ``\textit{The speech provides strong arguments}'' have a more local scope (specific arguments or points in the speech). Table \ref{tab:top_key_points} in Appendix \ref{sec:kpa_implementation} presents a more detailed list of pro and con key points, providing further insight into the evaluation criteria employed by the judge. We observe that less frequent key points reflect more nuanced aspects --- including remarks like ``\textit{The use of evidence shows thorough preparation}'' and ``\textit{The points are not well-developed}''.

In addition, the key points indicate that GPT-4.1 speeches are the most coherent and clear. This is unsurprising, considering how the speeches were constructed. For example, Arg-Human1 speeches consist of automatically concatenated arguments supporting the topic, while human speeches are \textit{transcribed from recordings}. \revision{The speech examples we provide in Appendix \ref{sec:speech_examples} illustrate the far more constructed and organized nature of}

GPT-4.1 speeches. The most notable difference is the division into paragraphs and a clear structure, including an opening, three arguments, and a summary. Given that LLM judges are notoriously 

biased towards stylistic properties \cite{wu2023style}, this may be one of the reasons that speeches by GPT-4.1 receive higher scores.

\section{Discussion}
In this work, we set out to test LLMs on debate speech evaluation, a challenging task that demands a nuanced understanding of long texts as well as a combination of multiple analysis and reasoning skills. Indeed, our findings show that this task is strictly the purview of larger (7B+ parameters) and more capable models. Moreover, we find that some state-of-the-art LLMs, such as o3, still underperform on this challenging evaluation task. 

The comparison of LLM judges to high-quality human annotations reveals some complex patterns. On the one hand, some LLMs are on par with individual humans in terms of instance-level agreement, occasionally even surpassing the performance of the typical human annotator. On the other, our deeper analysis demonstrates that all judges differ starkly from humans in their judgement behaviors -- LLM judges assign lower evaluation scores overall, and draw sharper distinctions between ``strong'' and ``weak'' speeches, potentially exaggerating performance gaps in ways that differ from human annotators. Moreover, judges still fail to accurately replicate the system-level ranking of speech sources given by humans.
These patterns echo recent works on the distinctions between system-level and instance-level judgment performance~\cite{gera2024justrank, gao2024re}.
Importantly, they point to open questions around the design of LLM-as-a-Judge evaluation pipelines and whether current methods adequately reflect nuanced human judgments.

Our results in Section \ref{sec:new_speeches_eval} indicate that modern LLMs can outperform humans in generating high-quality speeches. This raises concerns about misuse, particularly in settings where persuasive language could be weaponized by malicious actors. At the same time, it highlights promising applications in writing support or educational tools.

\paragraph{Future work} Improving the judging capabilities of smaller models remains an important goal\revision{, as this would expand accessibility and reduce the computational cost of deploying LLM-as-a-Judge systems. Beyond model size,} further research is needed to better align LLM judgment behavior with human evaluations, particularly in capturing nuanced patterns and system-level preferences. 

\revision{Another promising direction is to explore \emph{multi-dimensional} evaluations, where judges provide separate ratings for aspects such as rhetorical style, coherence, factual grounding, and persuasiveness, complementing the holistic score. Our preliminary use of Key Point Analysis (Section~\ref{sec:reasoning_analysis}) illustrates the potential of this approach and shows how richer frameworks can surface more fine-grained distinctions in model behavior.}

\revision{Future work may also examine the role of prompt design more systematically. Here we focused on a controlled zero-shot setting to isolate inherent judgment abilities. However, in-context learning (ICL) setups, where few-shot demonstrations are included, may substantially impact performance, particularly for weaker models. A broader comparison between zero-shot, CoT, and ICL conditions could help disentangle 
genuine judging capabilities from
adaptation to prompt structure.}

\revision{Finally, future work may investigate temporal alignment between training data, human annotations, and generated content. The speeches in our dataset are from 2020 or earlier, whereas modern LLMs are trained on more recent corpora. Although our topics are timeless, many domains evolve quickly, raising the question of whether improved speech quality reflects better reasoning, updated knowledge, or both. Systematic studies of knowledge and discourse shifts over time are needed to answer such questions.}

\section{Conclusion}

We present a novel benchmarking task for the LLM-as-a-Judge paradigm: rating long debate speeches arguing for or against a controversial topic. We conduct the first large-scale evaluation of judges in this challenging task and reveal a nuanced picture. While larger LLMs often align with human ratings at the instance level, they exhibit a tendency to be more critical, especially toward lower-quality speeches. 
Our findings also reveal that
modern LLMs may surpass humans in argumentative writing — a result that underscores both the impressive capabilities and the potential risks of such systems.

\section*{Limitations}\label{sec:limit}
\paragraph{Data-related limitations} In this work, we benchmark LLMs using data collected by \citet{slonim2021autonomous}. This imposes some limitations on our experiments. First, multiple quality dimensions, such as relevance, style, factuality, and argument strength, are all combined into a single score. This may introduce interpretability challenges, which we aim to address in Section \ref{sec:reasoning_analysis}. Second, the data by \citet{slonim2021autonomous} does not contain annotations for speeches by current SOTA models. 

We try to address this gap in Section \ref{sec:new_speeches_eval} by generating additional speeches using the state-of-the-art GPT-4.1. Moreover, analyzing how LLM judges rate synthetic outputs from older \textit{and} newer models offers valuable insight into the qualities they prioritize in a speech, as seen in Section \ref{sec:reasoning_analysis}.

Finally, in line with the data, we focus solely on opening debate speeches. This neglects the complexities of evaluating subsequent turns in debate, where arguments and counterarguments build upon each other. Evaluating LLMs' ability to judge multi-turn dynamics remains an open challenge.  

\paragraph{Variety of tested judges} In this work, we focus exclusively on prompted judges, as they allow for the use of complex, instruction-like prompts that closely resemble those given to human annotators \cite{slonim2021autonomous}. Evaluating the performance of fine-tuned judges---such as Prometheus \cite{Kim2024Prometheus2A} and LMUnit \cite{saadfalcon2024lmunitfinegrainedevaluationnatural}---on the task of rating debate speeches is left for future work. Reward models are trained for relative preference and are therefore unsuitable for the settings examined in this work.

\paragraph{SOTA speeches evaluation} In Section \ref{sec:new_speeches_eval}, we use five high-performing judges to analyze speeches authored by GPT-4.1. While our experiments show these models align well with human judges, we acknowledge that actual human annotation of this specific data could yield different results. Specifically, as we point out in Section \ref{sec:reasoning_analysis}, stylistic bias \cite{wu2023style} may have contributed to the higher score of these speeches. In addition, judges from the GPT-4.1 series may be biased toward self-generated texts \cite{xu2024prideprejudicellmamplifies}.

\paragraph{Prompt sensitivity}\revision{While we used the same prompt structure across models to maintain a controlled comparison setting, we recognize that this can disadvantage some models and benefit others. Our choice reflects a trade-off between per-model optimization and fair benchmarking. Importantly, optimizing prompts separately for each model risks entangling model-specific tuning with the evaluation itself, making it harder to draw generalizable conclusions. We acknowledge, however, that examining the sensitivity of results to alternative prompts is valuable, and we leave a systematic investigation of this aspect for future work.}

\paragraph{Inference reproducibility}
\revision{For efficiency, we ran models larger than $8$B parameters using 4-bit quantization, as full-precision inference was not feasible under our resource constraints. While quantization may slightly affect performance, we applied it consistently across all relevant models to ensure fairness.} 

\section*{Ethics Statement}
We make our data and code publicly available to ensure long-term reproducibility and enable others to build upon our work. The released speeches were collected by \citet{slonim2021autonomous}, which examines ethical concerns regarding the data annotation. \revision{We note that some debate topics in this dataset involve sensitive issues, and certain speeches could be considered offensive to some readers.}

Our analysis \revision{also} reveals potential risks of LLM misuse, though we recognize that studying these risks poses its own ethical challenges. For example, our findings could be used to identify models that excel at generating persuasive text beyond human capabilities and misuse them. This underscores the critical need for robust system safeguards and ethical deployment guidelines to prevent dangerous use.

\section*{Acknowledgments}
This research was conducted in collaboration between The Hebrew University of Jerusalem and IBM Research, and was partially supported by IBM-The Hebrew University Research Collaboration. We gratefully acknowledge the contributions and support of both institutions in enabling and facilitating this work.

\bibliography{custom}

\appendix

\section{Inference Implementation Details}\label{sec:judge-implem-details}
\begin{figure}[!htbp]
    \centering
    \includegraphics[width=\columnwidth]{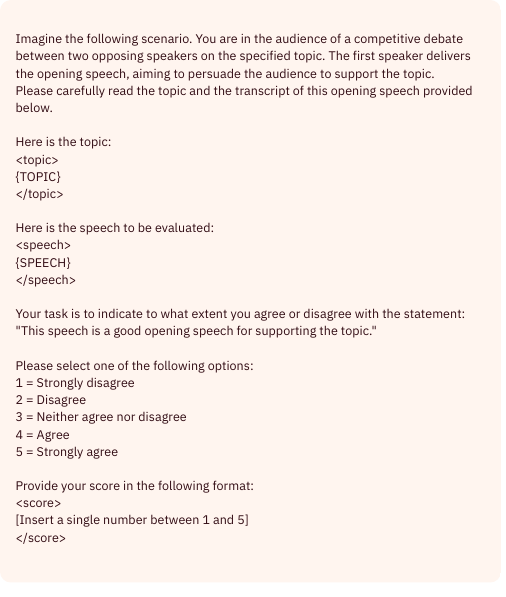}
    \caption{Speech scoring prompt. We give LLM judges a speech and its topic, and instruct them to rate it. The instructions closely follow the ones given to human annotators in \citet{slonim2021autonomous}.}
    \label{fig:zero-shot-prompt}
\end{figure}

\begin{figure}[!htbp]
    \centering
    \includegraphics[width=\columnwidth]{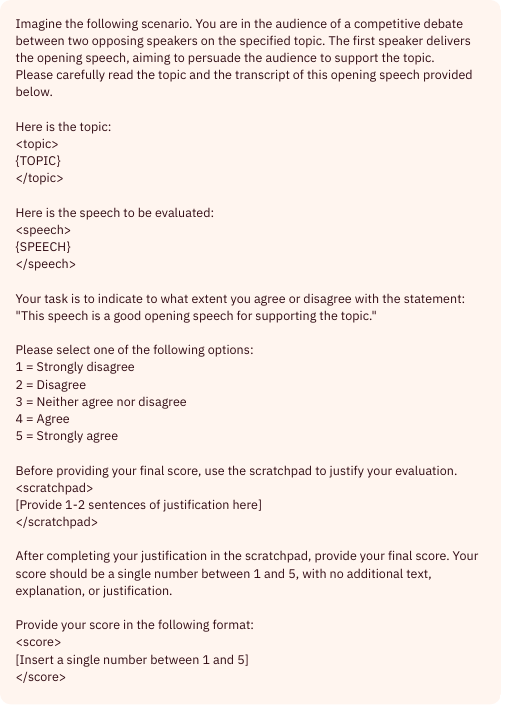}
    \caption{Chain of Thought (CoT) speech scoring prompt. We request the LLM judge to give a short justification for the given score.}
    \label{fig:cot-shot-prompt}
\end{figure}

Figure \ref{fig:zero-shot-prompt} presents the prompt used for the zero-shot experiment, while Figure \ref{fig:cot-shot-prompt} presents its Chain of Thought (CoT) variation. In all experiments, we run Llama and Qwen models larger than $8$B parameters using a quantized implementation ($4$bit). The temperature for all baselines is $0.01$ and the maximal context length is $4096$. \revision{Our experiments present results for running each judge model once.}

\paragraph{Runs variability}
\revision{We use a low temperature to promote deterministic behavior across runs. To verify this assumption, we conducted a small-scale replication with GPT-4.1, rerunning each speech three times under the same prompt. Both Tau and Kappa agreement scores remained stable, with variations no larger than $\pm{0.003}$ across runs, indicating high consistency.}

\section{Speech Generation Implementation}\label{sec:speech_gen_implementation}
\begin{figure}[!htbp]
    \centering
    \includegraphics[width=\columnwidth]{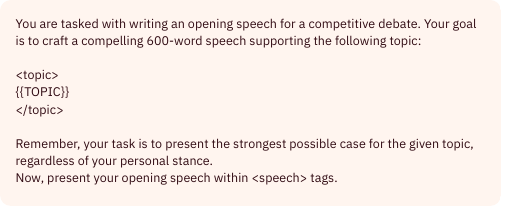}
    \caption{Speech generation prompt, requesting the model to generate a 600-word speech supporting a certain topic (e.g., "\textit{We should abandon social media}").}
    \label{fig:speech_gen_prompt}
\end{figure}
Figure \ref{fig:speech_gen_prompt} presents the prompt we use for speech generation. It requests the generation model to produce a 600-word speech supporting a specific debate topic (e.g., ``\textit{We should subsidize higher education}''). We use GPT-4.1 for the generation process, and create two speeches for each topic (similarly to human-authored speeches). For this step, we select a temperature of 1 and a maximal context of 1024. In case a generated speech exceeds the allowed 600 words, we iteratively remove sentences from its end until it complies with the required length.

\section{Additional Benchmark Details} \label{sec:additional_bencmark_details}
We take the benchmarking data in this study from \citet{slonim2021autonomous}. \revision{In this section, we add information on its collection methods (\S\ref{sub:og_data_collection}), and elaborate on the partition used in this study (\S\ref{sub:our_data_partition}). For complete details, please refer to the supplementary information from their work\footnote{\url{https://www.nature.com/articles/s41586-021-03215-w\#Sec14}}.}   

\subsection{Original Data Collection}\label{sub:og_data_collection}
\paragraph{Synthetic speech sources}
The following is a detailed review of the automatic pipelines used for speech generation:
\begin{itemize}
    \item \textbf{Summit} An extractive multi-document summarization system \cite{feigenblat2017unsupervised}. Given a topic, it selects relevant sentences and constructs a summary to serve as the speech.
    
    \item \textbf{Arg-Search} Based on the ArgumenText project \cite{stab2018argumentext}, which provides APIs for various argument mining tasks. Speeches are created by retrieving and combining arguments from ArgumenText in response to queries about a debate topic.
    
    \item \textbf{Speech-GPT2} Speeches generated by a GPT-2 model \cite{radford2019language} fine-tuned on a dataset of speeches \cite{orbach2020echochamberdetectingcountering}.
    
    \item \textbf{Arg-GPT2} A GPT-2 model \citep{radford2019language} fine-tuned on a dataset of arguments from \citet{gretz2020large}. Speeches are formed by automatically concatenating arguments generated by the fine-tuned model.
    
    \item \textbf{Arg-Human1}, \textbf{Arg-Human2} Speeches composed of automatically concatenated, crowd-sourced arguments on debate topics featured in the data. Arguments are sourced from \citet{gretz2020large} for Arg-Human1 and \citet{ein2020corpus} for Arg-Human2.
    
    \item \textbf{Project Debater} A system developed by IBM to compete in debates against humans \citep{slonim2021autonomous}. It processes large text collections, extracts relevant arguments, and generates speeches on a given topic.
\end{itemize}

\paragraph{Human-expert speeches} \revision{The dataset introduced by \citet{slonim2021autonomous} also contains speeches produced by human experts. These were transcribed from recordings of professional debaters delivering speeches in real time (see \citet{orbach2020echochamberdetectingcountering, mirkin-etal-2018-recorded}). For each topic, the dataset includes two such speeches, each recorded by a different debater.}

\paragraph{Annotation task} \revision{\citet{slonim2021autonomous} relied on crowd annotators to annotate speeches across sources. The annotators received the following instructions: ``\textit{Imagine the following scenario. You are in the audience of a competitive debate between two opposing speakers on the specified topic. The first speaker delivers the opening speech, aiming to persuade the audience to support the topic. Please carefully read the transcript of this opening speech provided below. For each of the following statements, please indicate to what extent you agree or disagree with the statements}''. They were then asked to indicate the extent to which they agreed with two statements: (1) ``\textit{This speech is a good opening speech for supporting the topic}'', (2) ``\textit{Most arguments in this speech support the topic}''. Responses were given on a five-point Likert scale, from Strongly disagree (1) to Strongly agree (5). Importantly, annotators were blind to the origin of the speech.}

\paragraph{Annotation quality} \revision{Inter-annotator agreement for the main question (``\textit{This speech is a good opening speech for supporting the topic}'') was $\kappa = 0.24$, which \citet{slonim2021autonomous} attribute to the task’s subjective nature rather than poor annotation quality. To validate reliability, they conducted three complementary checks:}  
\begin{enumerate}
    \item \textbf{Expert speeches:} \revision{Speeches delivered by human debate experts received consistently higher average scores than those produced by automatic systems.}
    \item \textbf{Control questions:} \revision{Annotators were also asked to rate a secondary statement --- ``\textit{Most arguments in this speech support the topic}'' --- which served as a control for annotator reliability.  In addition, ``control speeches'' of intentionally low quality were included to identify inattentive annotators.} 
    \item \textbf{Manual review:} \revision{A qualitative inspection of 20 speeches (10 high- and 10 low-scoring) confirmed that high scores corresponded to coherent, on-topic content, while low scores reflected off-topic, repetitive, or non-argumentative text.}
\end{enumerate}

\revision{This vetting process yielded $15$ human annotations per speech.}

\subsection{Data Partition Used}\label{sub:our_data_partition}
We use ``Pipeline-set-1'', a subset containing the most human-authored speeches. We removed 78 speeches that were used for vetting the annotation process, as these were designed to receive lower scores and identify poor quality annotations. We also excluded speeches from two topics --- ``\textit{We should increase airport racial profiling in the United States}'' and ``\textit{We should subsidize the human mission to Mars}'' --- since they lacked speeches from both Arg-Human1 and Arg-Human2.

\section{Parsing Errors}\label{sec:parsing_errors}
\begin{figure}[!htbp]
    \centering
    \includegraphics[width=\columnwidth]{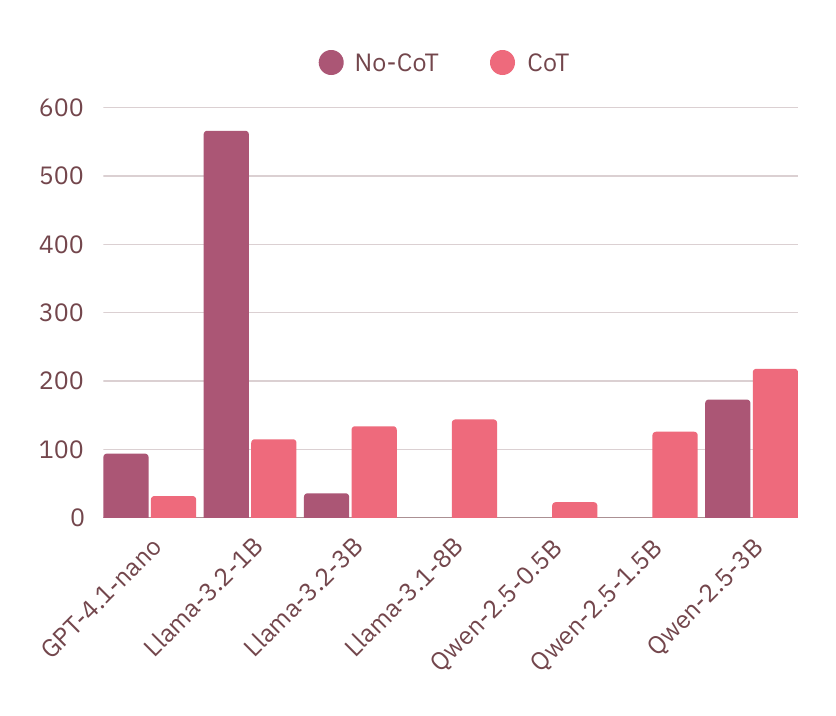}
    \caption{Number of parse errors for different prompt variations. ``No-CoT'' refers to the prompt described by Figure \ref{fig:zero-shot-prompt} and ``CoT'' to the prompt described by Figure \ref{fig:cot-shot-prompt}. We note that parsing errors mostly occur with smaller models and could largely vary with the prompt. In general, results for the CoT prompt seem to be more challenging to parse.}
    \label{fig:parsability}
\end{figure}

When the judge's response cannot be parsed, we assign the relevant speech a score of -1. Figure \ref{fig:parsability} presents the number of parsing errors for both prompt variations. We exclude models with fewer than 10 parsing errors for better readability. We observe that CoT prompting introduces parsing issues for some of the smaller models. The only exceptions are Llama-3.2-1B and GPT-4.1-nano, whose parsability is significantly improved by using CoT.

\section{Judge Ensembling} \label{sec:judge_ensem}
\begin{figure}[!htb]
    \centering
    \begin{subfigure}[b]{\linewidth}
        \centering
        \includegraphics[width=0.6\linewidth]{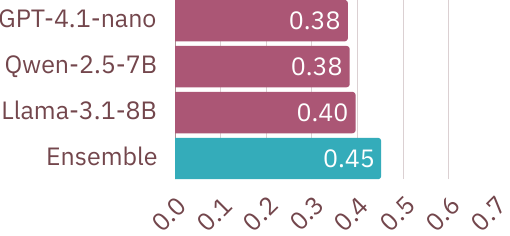}
        \caption{Weak judges ensemble (tau-c)}
        \label{fig:weak_ensemble}
    \end{subfigure}
    
    \vspace{2em}
    
    \begin{subfigure}[b]{\linewidth}
        \centering
        \includegraphics[width=0.7\linewidth]{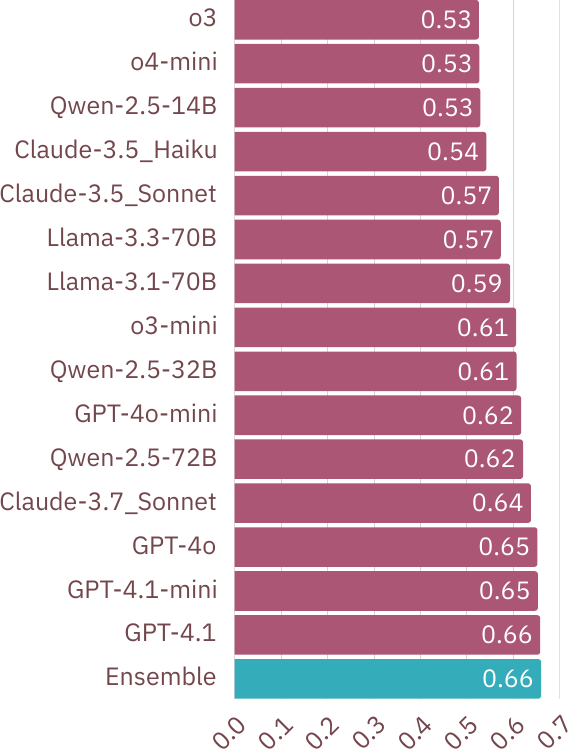}
        \caption{Strong judges ensemble (tau-c)}
        \label{fig:strong_ensemble}
    \end{subfigure}
    \caption{Results for two judge ensembles and their individual parts: Figure \ref{fig:weak_ensemble} shows an ensemble of relatively weak models (with $0.1<tau<0.5$) and Figure \ref{fig:strong_ensemble} shows a larger ensemble of more capable judges (with $tau>0.5$). Our results indicate that ensembling judges is a non-trivial enhancement technique, and might be beneficial only in certain cases.}\label{fig:ensembles_res}
\end{figure}

We inspect ensembling - aggregating the results of multiple diverse judges - as a means to improve LLM rating on this task. We examine two judge ensembles, divided according to their performance in the experiment presented in Figure \ref{fig:human_model_tau-c}. \textbf{(1) Weak judges}: all models with $0.1<tau<0.5$ \textbf{(2) Strong judges}: all models with $tau>0.5$. Figure \ref{fig:ensembles_res} presents results for both ensembles. We observe that ensembling the set of relatively weak models (GPT-4.1-nano, Qwen-2.5-7B, and Llama-3.1-8B) offers a slight improvement of 5 points. Interestingly, ensembling models with $tau>0.5$ has no effect. These findings demonstrate that model ensembling is a complex technique which can be beneficial in some scenarios but not in others.

\section{Additional Judges Scores Distributions} \label{sec:full_scores_dist}
\begin{figure}[!htbp]
    \centering
    \includegraphics[width=0.9\columnwidth]{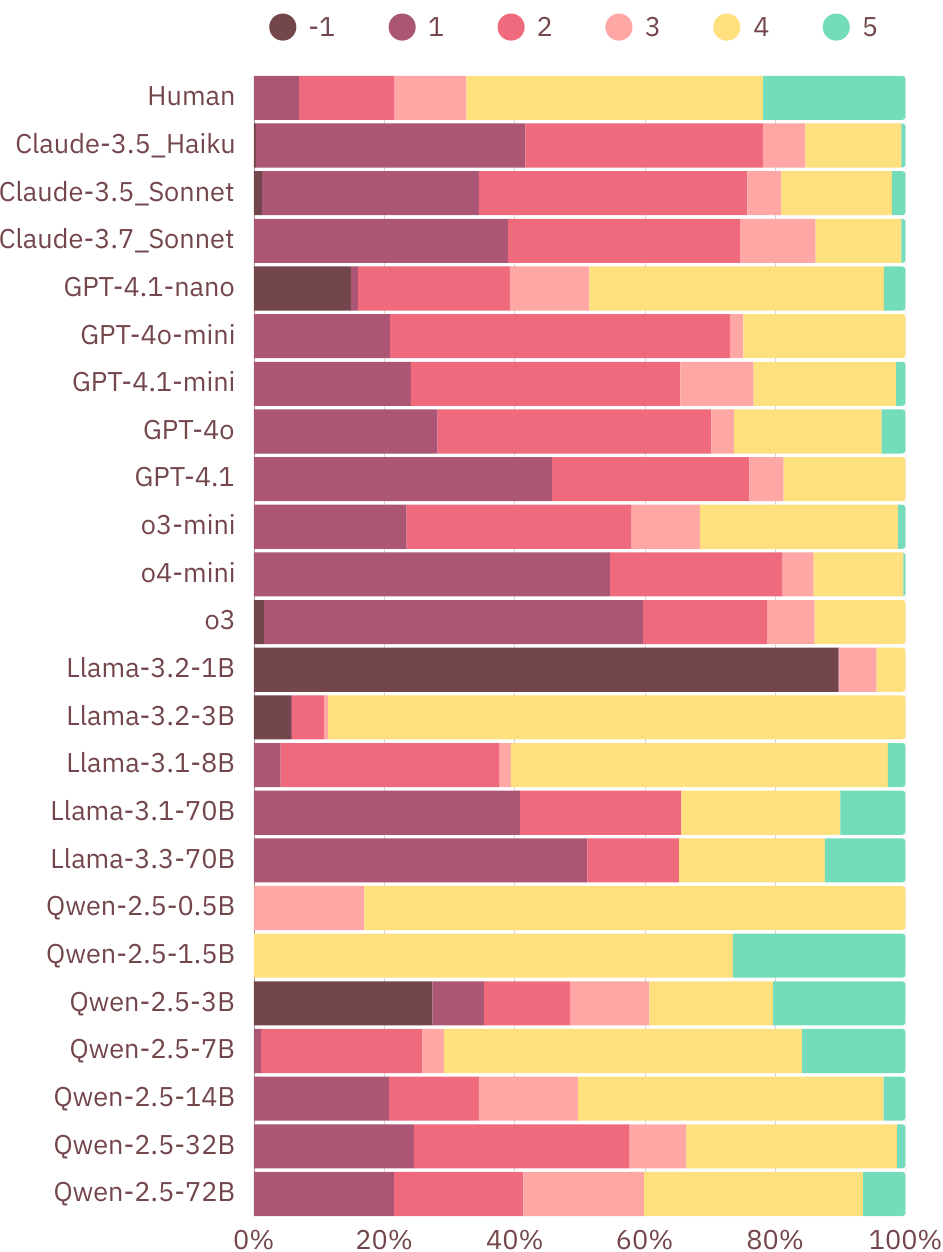}
    \caption{Judge scores distributions. Strong LLM judges tend to give lower scores than humans (-1 signifies parsing issues).}
    \label{fig:full_scores_dist}
\end{figure}

In Section \ref{sec:behavior}, we discuss the score distribution for a subset of the assessed judges. The same analysis for all judges is provided in Figure \ref{fig:full_scores_dist}.

\section{KPA: Additional Details}\label{sec:kpa_implementation}
\begin{figure}[!htb]
    \centering
    \includegraphics[width=\columnwidth]{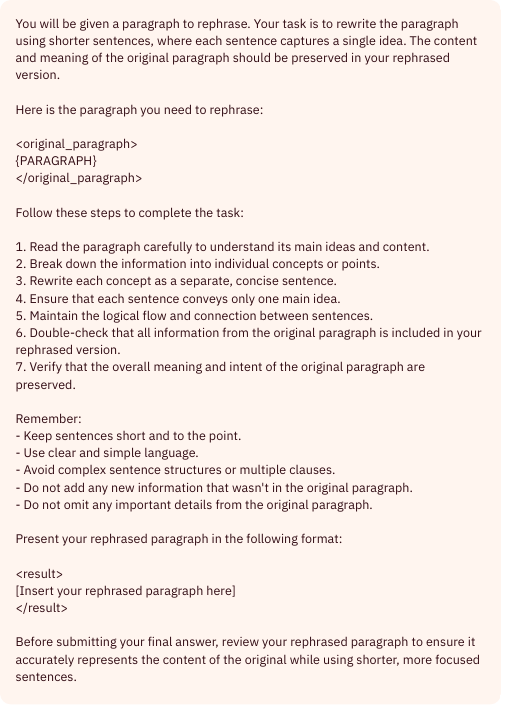}
    \caption{Key point preprocessing prompt: To improve key-point analysis, we convert chain-of-thought reasoning into shorter, clearer sentences that justify the speech score.}
    \label{fig:kpa_preprocess_prompt}
\end{figure}

\begin{table*}[!htb]
    \centering
    \begin{tabular}{ll}
    \toprule
        \cellcolor{lightgreen}\textbf{Pro key points} & \cellcolor{lightred}\textbf{Con key points} \\ \midrule
        \cellcolor{lightgreen}The speech provides strong arguments. & \cellcolor{lightred}The speech lacks a clear and focused argument.\\
        \cellcolor{lightgreen}The content supports the topic.& \cellcolor{lightred}It lacks persuasive reasoning.\\
        \cellcolor{lightgreen}Overall, the speech is a coherent opening.& \cellcolor{lightred}The speaker's points are often confusing.\\
        \cellcolor{lightgreen}The argument for reform is strong.& \cellcolor{lightred}The delivery is somewhat disorganized.\\
        \cellcolor{lightgreen}It presents a clear stance.& \cellcolor{lightred}There is no cohesive narrative.\\
        \cellcolor{lightgreen}It is well-structured.& \cellcolor{lightred}There is no clear thesis statement.\\
        \cellcolor{lightgreen}The delivery is persuasive. & \cellcolor{lightred}The points are not well-developed.\\
        \cellcolor{lightgreen}The use of evidence shows thorough preparation.& \cellcolor{lightred}The tone of the speech is inconsistent.\\
        \cellcolor{lightgreen}It clearly explains the benefits of socialism.& \cellcolor{lightred}The arguments are somewhat repetitive.\\
        \cellcolor{lightgreen}The speech stays focused on the topic.& \cellcolor{lightred}The speaker does not give a strong opening.\\
    \bottomrule
    \end{tabular}
    \caption{Top ten \green{pro} and \red{con} key points.}
    \label{tab:top_key_points}
\end{table*}

\begin{figure*}[!htb]
    \centering
    \includegraphics[width=\textwidth]{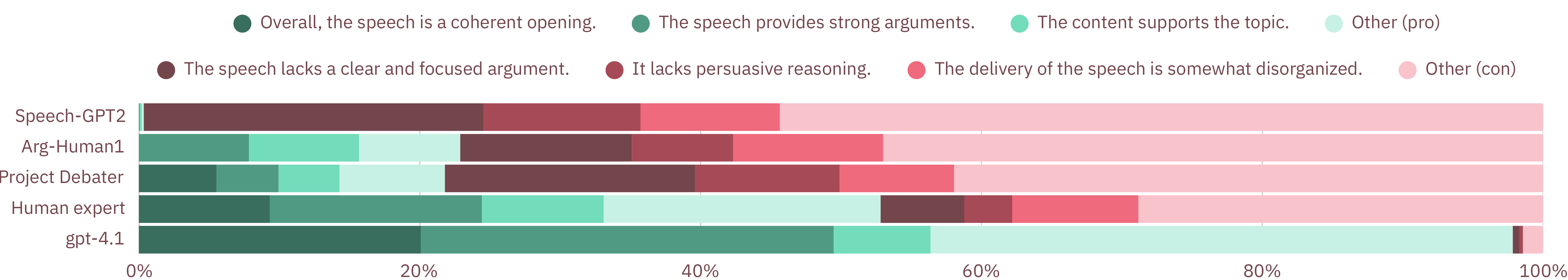}
    \caption{Distribution of the top three \green{pro} and \red{con} key points in \textbf{GPT-4.1-mini}’s chain-of-thought explanations, grouped by source. \green{Other (pro)} and \red{Other (con)} denote less frequent points. The relative share of positive key points per source reflects the judge ratings discussed in Sections~\ref{sec:source_analysis} and~\ref{sec:new_speeches_eval}}
    \label{fig:gpt-41-mini-kpa}
\end{figure*}

\begin{figure*}[!htb]
    \centering
    \includegraphics[width=\textwidth]{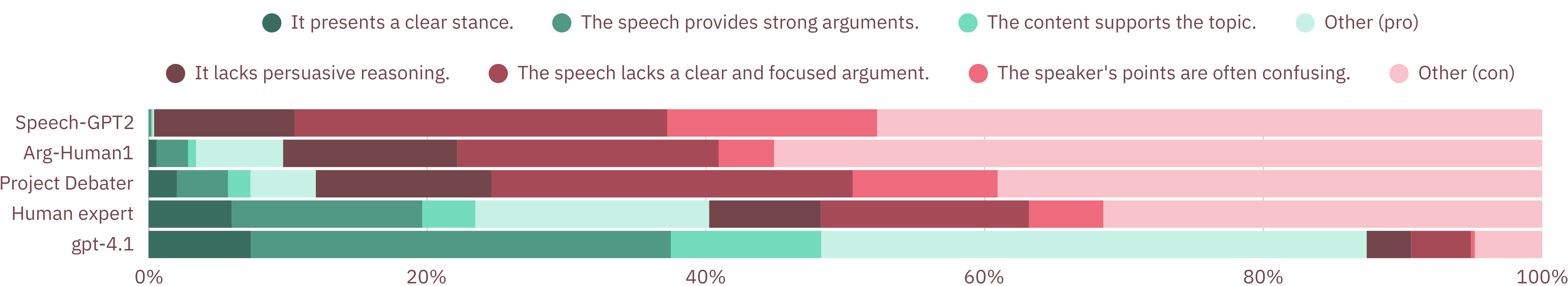}
    \caption{Distribution of the top three \green{pro} and \red{con} key points in \textbf{Qwen-32B}’s chain-of-thought explanations, grouped by source. \green{Other (pro)} and \red{Other (con)} denote less frequent points. The relative share of positive key points per source reflects the judge ratings discussed in Sections~\ref{sec:source_analysis} and~\ref{sec:new_speeches_eval}}
    \label{fig:qwen_reasoning_analysis}
\end{figure*}

In Section \ref{sec:reasoning_analysis}, we describe how we use Key Point Analysis (KPA) to interpret scores assigned by one of our strongest identified judges, Llama-3.3-70B, with CoT prompting. The analysis is designed for texts with relatively short sentences, such as arguments or product reviews. To better support this analysis, we use GPT-4.1 to rephrase the raw CoT strings into paragraphs of shorter sentences. Figure \ref{fig:kpa_preprocess_prompt} shows the prompt for this step. For a more robust analysis, we build the corpus using CoT reasoning from the three strongest CoT judges per model family, identified in \ref{sec:human-model_agreement}: Qwen-32B (CoT), Llama-3.3-70B (CoT), and GPT-4.1-mini (CoT). 

We report results for Llama-3.3-70B (CoT) in Section \ref{sec:reasoning_analysis}. Figures \ref{fig:gpt-41-mini-kpa} and \ref{fig:qwen_reasoning_analysis} report similar results for GPT-4.1-mini and Qwen-32B, respectively. 
\paragraph{Additional key points} Table~\ref{tab:top_key_points} presents the top 10 pro and con key points identified in our analysis, offering a more nuanced view of the judges' reasoning. Note that due to the extractive nature of KPA, some key points may be overly specific (e.g., “\textit{The argument for reform is strong}”).

\section{Speech Examples}\label{sec:speech_examples}
\begin{figure*}[!htb]
    \centering
    \begin{tcolorbox}[colback=mustard!5!white,colframe=mustard!75!black]
    ``We should increase fuel tax.  So, in this debate, we're going to argue that the states should introduce a tax on fuels, on fuels, in order to recoup the cost of fuel.  So, in the status quo, states are allowed to levy a small tax on every single product that they consume.  But they can't levy a tax on fuels, because that would be an unfair tax on people who are worse off in the world.  We think that the state should instead tax fuels, and if the cost of fuel exceeds the cost of fuel, the state should tax them.  So, with that in mind, we have three arguments for why we should tax the fuel.  The first is that we think that this will be a fair and just solution for the climate crisis.  So, right now, there's a huge debate in the developed world about climate change.  The most important question, of course, is whether we can stop global warming dead in its tracks, and the answer is no.  We need to move as quickly as possible to reduce the amount of carbon that is put into the atmosphere, which will inevitably lead to global warming, which will lead to things like rising sea levels and more extreme weather.  The only way to do this, therefore, is to put as many people as possible in power who can push the buttons, push the levers, and turn the lights off.  But we think that, right now, there's a huge power imbalance in the developed and developing world, and that, ultimately, the only way to make a change, to fight climate change, is to have a market that is regulated, that is controlled, that is controlled by the people who live on the ground.  We think that this is the only way to do it, because the people who live on the ground are the most informed, and the people who are making the decisions are the most educated.  We think that, therefore, the people who are most affected by climate change are the people on the ground, and that we need to put the people who are most affected in power.  The reason for this is that, right now, the average person in the developed world is only a single vote out of a whole lot of people in the developed world.  We think that, therefore, if you put a tax on the people living on the ground, that they're more likely to take action, and that this is a one-time opportunity for people to change their behavior.  The people who are in power right now are very very different from the people who are in power right now, and it's clear that, in the future, people will have more power, and we'll be able to use that power more efficiently.  Our second argument is that this will save states money.  So, right now, if states want to build up their energy infrastructure, they're spending a lot of money to do so.  And that's a problem.  We see that, for instance, in the US, that a big oil companies are using up all of the oil in the country, and that's why the price of oil is going up.  It's a problem that, right now, the US is spending more than \$100 billion a year on oil.  But, right now, we're spending less than \$1 trillion a year on oil, which is a fraction of what we need.  We think that if you tax the energy that the state uses, that that money will go much further.  For instance, if you have a big nuclear power plant, and the price of nuclear power goes up, and people are using that for their own energy, that means that you're going to be able to get more of the energy that you need for your own energy needs, and that money will go further.  And, in the same way, a tax on fuel will also reduce the cost of fuel.  So, ultimately, because you will be able to recoup the cost of fuel, and because this will save the state money in the long term, we think that this is a good idea.  For all these reasons, we should tax fuel.  Thank you.''
    \end{tcolorbox}
    \caption{\textbf{Speech-GPT2} speech on the topic ``\textit{We should increase fuel tax}''}
    \label{fig:gpt2_speech}
\end{figure*}

\begin{figure*}[!htb]
    \centering
    \begin{tcolorbox}[colback=mustard!5!white,colframe=mustard!75!black]
    ``We should increase the fuel tax. Fuel in the united states is very cheap relatively to other developed nations in the world and so is the tax rate on gas. This contributes to the existing situation in which americans end up consuming more gasoline than nearly any other country. This creates a number of problems two of which we shall discuss in this speech. A, a high level of pollution that harms the environment. So the more that people are driving, the more it pollutes the environment. We think that's pretty elementary. This pollution of course is a negative externality that the entire society is forced to pay the price for. Raising the tax which makes fuel more expensive can be used in order to make consumers internalize this cost and this way we can reach a better equilibrium that takes into account the harm to society and to the environment. This tactic of negative taxation to deal with negative externalities is very common in various other fields. For instance if there is a factory that pollutes like the waters of a river nearby that there's a and there's like a village that is using that water from the river for drinking, then we have a market failure because the cost of pollution isn't felt by the factory owner and so this factory owner has an incentive to continue on polluting as much as as much as possible. In order to deal with this what we do is we make him pay a tax on pollution that is proportional to the amount of damage that is being caused as a result from the pollution. And so the factory will take care of reducing the amount of pollution until the like amount that is going to be optimal. Same thing exactly goes with drivers and fuel. Second thing is, on the second problem, is a high dependency on on private transportation that is presenting other alternatives from being formed. So seeing as gas is cheap, many people are incentivized to own a car. And so other alternatives to transportation just don't seem appealing enough. These alternatives can be things like, I don't know, a higher quality public transportation, carpool and so on. Why is this so bad? Because judging by other places in the world these alternatives are potentially cheaper, greener, and most importantly they will make the US less dependent on foreign oil. The biggest exporters of fossil fuels are countries that are not exactly friendly to the united states and the dependency on fuel forces us to pay a heavy political price. For instance, it's provides countries like saudi arabia with a very strong position of power that doesn't allow the US to criticize the horrible way in which women women's rights are being trampled by the local regime there. Raising the fuel tax can help us become more independent and that's great. Another advantage in raising the fuel tax is in lies in the ability to use this revenue from taxation in order to improve the status of infrastructure for roads and highways that desperately requires improvement. The budget for these type of things is based on incomes and revenues coming from these taxes exactly and that is why they are in such bad shape today. With so many drivers even like a one dollar raise per gallon means huge amounts of money that could be used to to even lower other types of taxes like income tax. So for all of these reasons please propose.''
    \end{tcolorbox}
    \caption{\textbf{Human-expert debater} speech on the topic ``\textit{We should increase fuel tax}''}
    \label{fig:human_speech}
\end{figure*}

\begin{figure*}[!htb]
    \centering
    \begin{tcolorbox}[colback=mustard!5!white,colframe=mustard!75!black]
    ``Ladies and gentlemen, esteemed judges, and fellow debaters,

Today, I stand in firm affirmation of the motion: "We should increase fuel tax." This is not just a matter of fiscal policy; it is a matter of economic prudence, environmental stewardship, and social responsibility. Allow me to lay out the compelling reasons why increasing fuel tax is not only justified, but necessary.

First and foremost, let us consider the environmental imperative. The combustion of fossil fuels is the primary driver of climate change. According to the Intergovernmental Panel on Climate Change, transportation accounts for nearly a quarter of global carbon dioxide emissions. Fuel taxes are a proven and practical tool to address this crisis. By making fossil fuels more expensive, we create a financial incentive for individuals and businesses to reduce consumption, seek alternatives, and invest in cleaner technologies. Countries like Sweden and Norway have shown that sustained, targeted fuel taxes can lead to significant reductions in greenhouse gas emissions while maintaining economic growth.

Secondly, increasing fuel tax directly combats the many hidden costs—what economists call externalities—associated with fossil fuel consumption. Every liter of gasoline burned not only pollutes the air but also inflicts damage on public health through increased rates of asthma, heart disease, and premature death. These are costs society collectively bears—in higher healthcare expenditures, lost productivity, and diminished quality of life. A higher fuel tax internalizes these costs, making polluters pay and shifting the burden away from taxpayers and vulnerable communities.

Third, a higher fuel tax can provide critical funding for infrastructure and public transportation. Throughout the world, roads, bridges, and transit systems are deteriorating due to chronic underfunding. By linking fuel consumption to infrastructure funding, we create a sustainable revenue stream. Take Germany, for example—revenue from fuel taxes is a cornerstone of their robust, high-quality public transit networks. Improved infrastructure not only benefits commuters, but also businesses, emergency services, and the economy as a whole.

Now, some may argue that a fuel tax is regressive and burdens lower-income households. This is a concern we take seriously. However, evidence shows that with thoughtful policy design, these impacts can be mitigated or even reversed. For example, a portion of fuel tax revenue can be redirected as targeted rebates or social benefits for low-income groups. Furthermore, investment in affordable public transportation offers everyone, regardless of income, greater mobility and access to opportunity. The principle is simple: we can, and must, use fuel tax revenue to correct past inequities, not deepen them.

Moreover, increasing fuel tax encourages innovation and economic diversification. When fuel is artificially cheap, there is little incentive to develop or adopt new technologies. By aligning the price at the pump with the true cost to society, we spur investment in electric vehicles, sustainable urban planning, and renewable energy—all sectors poised to create the jobs of the future. Look at how surging oil prices over past decades have unleashed waves of innovation in cleaner, more efficient vehicles. A fuel tax creates a stable, predictable path forward, allowing businesses to adapt and thrive.

Finally, addressing climate change and transitioning to sustainable energy is not a burden: it is an opportunity. Countries that act now—not later—will be the leaders of the 21st century economy. Rising to meet our environmental and fiscal challenges with courage and foresight is not just good policy; it is good leadership.

In summary, increasing the fuel tax is a powerful, evidence-based way to protect our climate, our health, our economy, and our future. It is a policy that, when designed with fairness and ambition, offers extraordinary benefits to both current and coming generations.

Thank you.''
    \end{tcolorbox}
    \caption{\textbf{GPT-4.1} speech on the topic ``\textit{We should increase fuel tax}''}
    \label{fig:gpt_41_speech}
\end{figure*}

Figures \ref{fig:gpt2_speech}, \ref{fig:human_speech}, \ref{fig:gpt_41_speech} provide examples of speeches from different sources: Speech-GPT2, Human expert, and GPT-4.1. All speeches discuss the same topic (``\textit{We should increase fuel tax}'').

\end{document}